\documentclass[runningheads]{llncs}

\usepackage{graphicx}
\usepackage{hyperref}
\usepackage{amsmath}
\usepackage{algorithmic}
\usepackage{algorithm}

\usepackage{makecell}
\usepackage{multirow}
\usepackage{threeparttable}
\usepackage{booktabs}

\usepackage{subfigure}
\begin{document}
%
\title{HeroNet: A Hybrid Retrieval-Generation Network for Conversational Bots}
%


\author{Bolin Zhang \inst{1}  \and Yunzhe Xu
\inst{1}  \and Zhiying Tu\inst{1} \thanks{Corresponding author.} and Dianhui Chu \inst{1}}

\authorrunning{Zhang and Xu et al.}


\institute{Harbin Institute of Technology, ICES Center \\ \email{brolin@hit.edu.cn}, \email{181310122@stu.hit.edu.cn} \\ \email{\{tzy\_hit, chudh\}@hit.edu.cn}\\}
\maketitle              
\begin{abstract}

Using natural language, Conversational Bot offers unprecedented ways to many challenges in areas such as information searching, item recommendation, and question answering. Existing bots are usually developed through retrieval-based or generative-based approaches, yet both of them have their own advantages and disadvantages. To assemble this two approaches, we propose a \textbf{h}ybrid r\textbf{e}t\textbf{r}ieval-generati\textbf{o}n \textbf{net}work (\emph{HeroNet}) with the three-fold ideas: 1). To produce high-quality sentence representations, \emph{HeroNet} performs multi-task learning on two subtasks: Similar Queries Discovery and Query-Response Matching. Specifically, the retrieval performance is improved while the model size is reduced by training two lightweight, task-specific adapter modules that share only one underlying T5-Encoder model. 2). By introducing adversarial training, \emph{HeroNet} is able to solve both retrieval\&generation tasks simultaneously while maximizing performance of each other. 3). The retrieval results are used as prior knowledge to improve the generation performance while the generative result are scored by the discriminator and their scores are integrated into the generator's cross-entropy loss function. The experimental results on a open dataset demonstrate the effectiveness of the \emph{HeroNet} and our code is available at \url{https://github.com/TempHero/HeroNet.git}

\end{abstract}

\section{Introduction}

By using natural language, a more direct form of interaction, conversational bot will play a ``bridge" role between a user and massive services, improving the user oriented service delivery\cite{esbot}. However, generating attractive and informative responses according to user queries is still a major obstacle to bot development. Conversational bots are usually built on either generation-based or retrieval-based models, but both of which have their own pros and cons\cite{REAT}. 

Since retrieval-based models aim to find the best matching result from a pool of real human responses, they will return to the user a fluent and informative sentence with diversity\cite{Rpros}. But the performance of these models is limited by the size and the quality of the response pool. On the other hand, generation-based models are capable of generalizing to unseen context and returning new responses not covered in the history pool. But their responses are likely to be very general or universal\footnote{ This means that the bot often responds with universal answers such as ``Thank you", ``I don't know", ``I have no idea".}, with insufficient information\cite{bolin}. Therefore, it is worthwhile to study how to assemble retrieval methods and generation methods, so as to make use of their strengths and avoid their weaknesses.

\begin{figure}[!h]
\includegraphics[width=\textwidth]{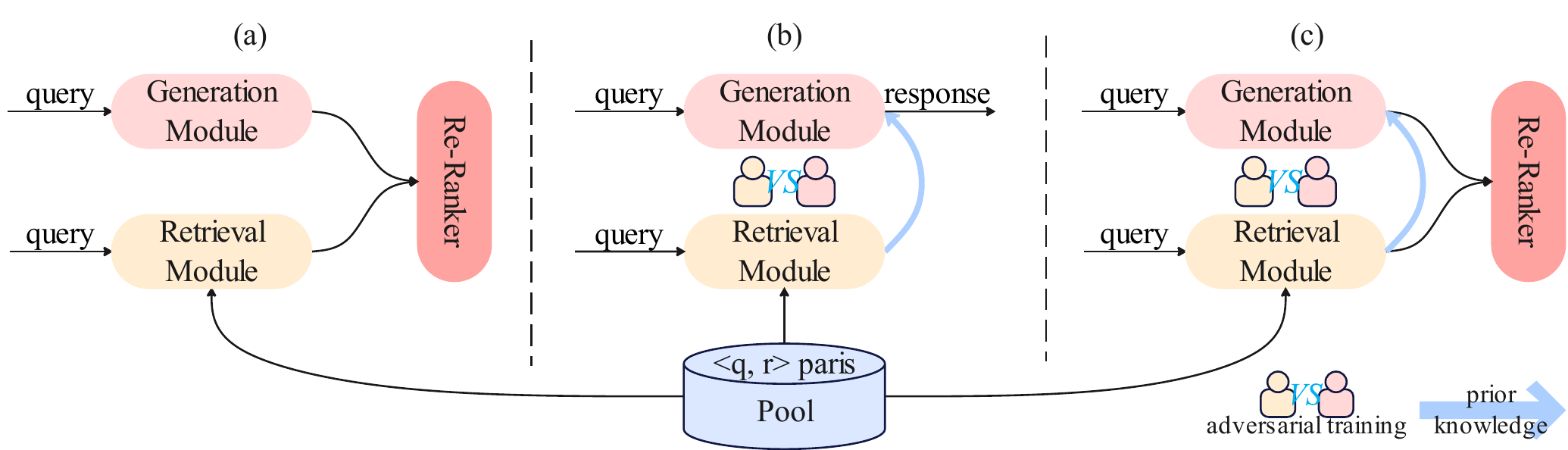}
\caption{Three common schemes for retrieval-generation ensemble: (a) Vanilla Re-Rank methods, (b) Retrieval-Enhanced Methods, (c) Mutual Gain Methods. Adversarial training aims to improve the performance of both the generation and retrieval modules by playing a min-max game between the two. Prior knowledge, which is the most relevant response extracted by the retrieval module, will be fed into the generation module to help it produce more informative sentences.} 
\label{intro}
\end{figure}

As shown in Fig.~\ref{intro}, the existing works could be grouped into three different categories: a)Vanilla Re-Rank methods, b) Retrieval-Enhanced methods, and  c) Mutual Gain methods. Given the conversation context, Vanilla Re-Rank methods \cite{ma1,ma2,ma3,HybridNCM} find $m$ related response via retrieval module and generate one response via generation module, and then reorder this $m$+$1$ results via re-ranker module. Since acquiring external structural knowledge is costly, Retrieval-Enhanced methods take $m$ most relevant responses ranked by the retrieval module as prior knowledge, and utilize them to help the generation module to produce a more informative response\cite{REAT,mb1,IRGAN}. Moreover, to improve retrieval and generation modules, generative adversarial networks are introduced. Vanilla Re-Rank methods focus on improving the performance of re-ranker modules, while  Retrieval-Enhanced methods focus on improving the performance of generation modules. Combining these two merits, Mutual Gain Methods\cite{EnsembleGAN} aim to find $m$ related responses via retrieval module and generate $n$ responses via generation module, and then reorder this $m$+$n$ results via re-ranker module. These type of methods usually contains multiple encoders and decoders with different functions, resulting in a complex model architecture.

To this end, we proposed a \textbf{h}ybrid r\textbf{e}t\textbf{r}ieval-generati\textbf{o}n \textbf{net}work (\emph{\textbf{HeroNet}}), which has a simple but effective architecture that leverages both prior knowledge and adversarial training. As shown in Fig.~\ref{heronet}, \emph{HeroNet} contains only one T5-Encoder and one T5-Decoder, which will undergo three learning processes: i) retrieval multi-task learning, ii) generation adversarial learning and iii) re-rank learning. 

In the retrieval multi-task learning, the T5-Encoder will be trained on the two subtasks (Similar Queries Discovery and Query-Response Matching) with two different task-specific adaptors ($\psi_D$ and $\psi_M$). By this way, the sentence embedding performance of the T5-Encoder will be improved and the model size will be reduced. In the generation adversarial learning, the trained T5-Encoder and $\psi_M$ will be regarded as the Discriminator $D_\phi$ and the T5-Decoder will be regarded as the Generator $G_\theta$. The policy gradient \cite{pg} is used to help $G_\theta$ generate $k$ responses and $D_\phi$ will judge which is the best. In the re-rank learning, the adversarial trained $D_\phi$ will rank the retrieved $m$ results and the generated $k$ results. 

The main contributions are summarized as follows:
\begin{itemize}
    \item To combines the merits of both retrieval-based generation-based methods, we propose a \textbf{h}ybrid r\textbf{e}t\textbf{r}ieval-generati\textbf{o}n \textbf{net}work (\emph{\textbf{HeroNet}}), which has a simple but effective architecture applied three process of learning.
    \item By introducing multi-task learning, the performance of sentence embedding is improved in \emph{HeroNet}. Moreover, the generation and retrieval performance of \emph{HeroNet} is also improved by exploiting adversarial training.
    \item Experimental results on open datasets \emph{UbuntuV2} demonstrate the effectiveness of \emph{HeroNet}. Codes are also publicly available at \url{https://github.com/TempHero/HeroNet.git}.
\end{itemize}

\section{Related Works}\label{Related Works}
The retrieval-generation ensemble methods have been attracting increasing attention in recent years. Multi-Seq2Seq \cite{ma1} employs $k+1$ encoders, one for query and other $k$ for retrieved results. The decoder receives the outputs of all encoders, and remains the same with traditional seq2seq for sentence generation. Then, GBDT is deployed to re-rank the retrieved results and generative results. REAT \cite{REAT} consists of a discriminator $D$ and a generator $G$, both of which are enhanced by N-best response candidates from Lucene\footnote{https://lucene.apache.org/}. EnsembleGAN \cite{EnsembleGAN} consists of a language-model-like generator, a ranker generator, and one ranker discriminator. The two generators aims to generate improved highly relevant responses and competitive unobserved candidates respectively, while the discriminator aims to identify true responses. HybridNCM \cite{HybridNCM} consists of three modules: i) generation module that employs a context encoder, a facts encoder and a response decoder, ii) retrieval module that employs Lucene, and iii) hybrid ranking module that employs CNN Layers and MLP. The architectures of these models are somewhat complex compared with our model (\emph{HeroNet}) with a simple but effective architecture. The details of these works' comparison are shown in Table.~\ref{comparison}

\begin{table}[htbp]
  \centering
  \caption{The Comparison of existing works}
  
    \begin{tabular}{cccc}
    \hline
    Method & \multicolumn{1}{l}{Num of Encoders} & \multicolumn{1}{l}{Num of Decoders} & Methods \\
    \hline
    Multi-Seq2Seq\cite{ma1} & k+1   & 1     & Fig.~\ref{intro}(a) \\
    HybridNCM\cite{HybridNCM} & 2     & 1     & Fig.~\ref{intro}(a) \\
    REAT\cite{REAT}  & 4     & 1     & Fig.~\ref{intro}(b) \\
    EnsembleGAN\cite{EnsembleGAN} & 1     & 2     & Fig.~\ref{intro}(c) \\
    HeroNet(\textbf{Ours}) & 1     & 1     & Fig.~\ref{intro}(c) \\
    \hline
    \end{tabular}%
  \label{comparison}%
\end{table}%

\section{Methods}\label{Methods}
\subsection{Overview of the Model}
\label{Overview}
\begin{figure}[!ht]
\includegraphics[width=0.85\textwidth]{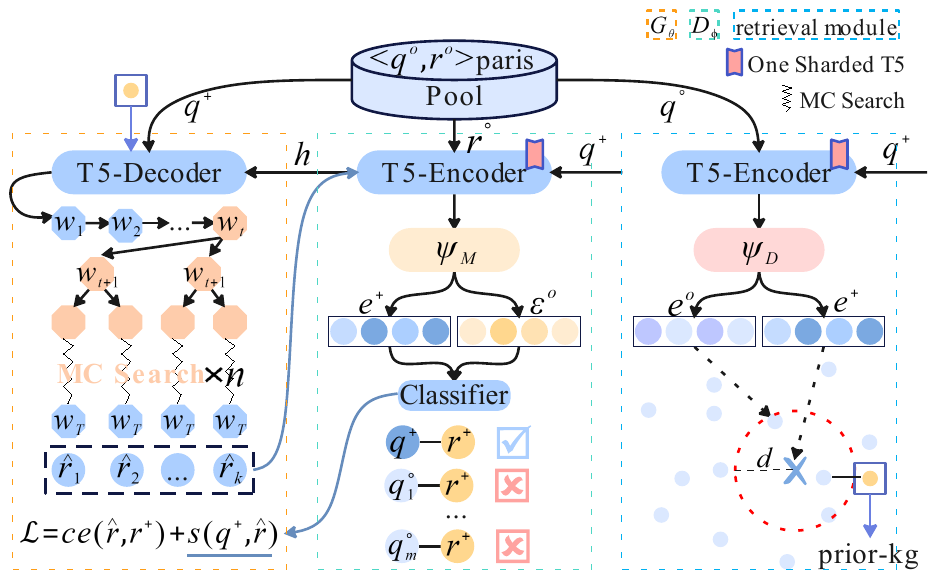}
\centering
\caption{The overview of \emph{HeroNet}, which consists of only one shared T5-Encoder and one T5-Decoder with the policy gradient and two adapters. The two encoders are the same things and share the same parameters. To make it easier to read, we have drawn them separately.}
\label{heronet}
\end{figure}

The architecture of \emph{HeroNet} is shown in Fig.~\ref{heronet}, which consists of only one shared T5-Encoder, one T5-Decoder with the policy gradient, and two task-specific adaptors ($\psi_M$ and $\psi_D$). Given a user query $q^{+}$, \emph{HeroNet} will be applied three learning processes: i) in Retrieval Multi-task Learning, it retrieves $m$ responses $\hat{r_i}$, $i=1,2..m$ from candidates pool $\langle q^{o},r^{o} \rangle $, ii) in Generation Adversarial Learning, it generates $n$ responses $\hat{r_j}$, $j=1,2..n$ based on Monte Carlo Search\cite{seqgan}, and iii) in Re-rank Learning, it ranks these $m$+$n$ responses leveraging the T5-Encoder trained in the process (i). The first of the re-ranked responses is taken as the generated result, and the first $k$ of the responses are taken as the retrieved result. \emph{HeroNet} utilize T5 \cite{T5} as the backbone, which is an advanced text generation Seq2Seq model and consists of an encoder and a decoder.

\subsection{Retrieval Multi-task Learning}

To improve the sentence embedding of T5-Encoder, we introduce two sub-tasks: Similar Queries Discovery and Query-Response Matching. Inspired by \cite{multi-dataset}, Hero-Net is aiming to combine the benefits of multi-task by learning a collection of task-specific adaptors that share an underlying T5-Encoder model. 

\subsubsection{Similar Queries Discovery (SQD) Task}: Given a query $q_{i}$ and the target number $m$, the goal is to discover $m$ queries $q_{i}^{o}$, $i=1,2..m$ from candidates pool $\langle q^{o},r^{o} \rangle $, which are semantically similar to $q_{i}$.

Followed by MEAN-strategy\cite{st5}, each query will be encode into a sentence embedding $e$ by averaging all token representations from the T5-Encoder. Then the fixed-length embedding vectors of $q_{i}$ and $q^{o}$ will be encoded as $e_i$ and $e^{o}$, respectively. After encoding, the SQD-specific adaptor $\psi_D$ applies the Projection Layer and Layer Normalization to $e_i$ and $e^{o}$. Thus, $q_{i}$ and $q^{o}$ are finally converted to $v_{i}$ and $v^o$ respectively. The operations can be can be represented as a function $f_{D}$. 
The distance between $v_i$ and $v^o$ are calculated as:

\begin{equation}
d(v_i, v^o) = \sqrt{(v_{i} - v^{o}) (v_{i} - v^{o})^T }
\label{distance}
\end{equation}
, and $v_{i}$ and $v^{o}$ are calculated as:

\begin{equation}
v_{i} = f_{D}(e_{i}) = W_D \times e_{i} + b_D 
\end{equation}

\begin{equation}
v^{o} = f_{D}(e^{o}) = W_D \times e^{o} + b_D 
\end{equation}
,where $W_D$ represents the trainable weight parameters of $\psi_D$ and $b_D$ is the trainable bias parameters of $\psi_D$. At last, the top $m$ queries $q_{j}^{o}$, $j=1,2..m$, will be extracted from $\langle q^{o},r^{o} \rangle $ pool according to the distance calculation in the Eq.~\ref{distance}. By the way, the negative samples are selected from the candidates pool using the traditional retrieval algorithm BM25, which lacks the consideration of semantic similarity.

\subsubsection{Query-Response Matching (QRM) Task}: Given a query $q^{+}$, the goal is to find the $r^{+}$ paired with $q^+$ from the candidates pool $\langle q^{o},r^{o} \rangle $. Each response $q^{o}$ from the pool will be calculated a matching score of how well it matches $q^+$.

After encoding operation of the shared T5-Encoder in SQD task, the $q^+$ and $r^{o}$ are converted to $e^+$ and $\varepsilon^{o}$ respectively. Then, the QRM-specific adaptor $\psi_M$ applies the Projection Layer and Layer Normalization to $e^+$ and $\varepsilon^{o}$. Followed by the similar operations in $\psi_D$, $q^+$ and $r^{o}$ are encoded to $p^+$ and $p^o$ finally. These operations can be can be represented as a function $f_{M}$. Thus, $p^+$ and $p^o$ are calculate as:

\begin{equation}
p^+ = f_{M}(e^+) \quad ; \quad p^o= f_{M}(\varepsilon^o) 
\end{equation}

They are concatenated with the element-wise difference $|p^+ - p^o|$ and multiply it with the trainable weight $W_{M}$ of $\psi_M$. Thus, the matching score between $q^+$ and $r^{o}$ will be calculated as:

\begin{equation}
\label{score}
s(q^+, r^{o})= \sigma(W_{M} \times concat(p^{+}, p^{o},|p^{+} - p^{o}|) )
\end{equation}

, where $\sigma$ represents the $sigmoid$ function and $|p^{+} - p^{o}|$ measures the distance between the dimensions of $p^{+}$ and $p^{o}$, ensuring that matching pairs are closer and ill-matching pairs are further apart. 

Instead of using random negative sampling, the retrieved queries from the SQD-Task by $\psi_D$ $q_{j}^{o}$, $j=1,2..m$ are used to build negative samples $\langle q_{j}^{o},r_{+} \rangle $. Moreover, the $m$ responses $r_{j}^{o}$, $j=1,2..m$ paired with the retrieved queries are also used to build negative samples $\langle q^{+},r_{i}^{o} \rangle $.

\subsection{Generation Adversarial Learning}
Inspired by the breakthrough of generative adversarial networks (GANs) on text generation\cite{LeakedGAN,MaskGAN,RelGAN}, the adversarial training is introduced improve jointly the performance of retrieval and generation modules. By calculating the matching scores of $\langle q^{+}, \hat{r_{i}} \rangle $ pairs, the discriminator $D_\phi$ aims to measure the quality of the generative responses. By employing a policy gradient and Monte Carlo search \cite{seqgan}, the generator $G_\theta$ aims to output $n$ responses $\hat{r_{i}}$, $i=1,2..n$ to fool the discriminator $D_\phi$. $D_\phi$ consists of the shared T5-Encoder and adapter $psi_M$ which are trained in the Retrieval Multi-task Learning.

\subsubsection{Global Optimization Objection}


Given a query-response pair $\langle q_{i},r_{i} \rangle $, the discriminator $D_{\phi}$ (T5-Encoder and Adaptor $\psi_M$) will extract $m$ negative responses from the candidates pool, so process can be formalized as $r \sim P_{-}(r^{-}|q_i)$. The generator $G_{\theta}$ will output $n$ responses $\hat{r}_{[1:n]}$ based on $q_{i}$, which should be regarded as negative samples too. Analogously, this process is formalized as $ \hat{r} \sim G_{\theta}(\hat{r}|q_i)$. Moreover, the positive samples construction is formalized as $r \sim P_{+}(r_i|q_i)$. Followed by the idea of GAN\cite{gan}, \emph{HeroNet} aims to unify these two different types of models ($G_\theta$ and $D_\phi$) by letting them play a mini-max game. The overall optimization Objection is defined as: 

\begin{equation}
\label{overall}
\begin{aligned}
J^{G^{*}, D^{*}}= &\min _{\theta} \max _{\phi} 
\sum_{i=1}^{N} \{ \mathrm{E}_{r \sim P_{+}(r_i|q_i)}[\log D_{\phi}(\varepsilon_{i}|e_{i})] \\
+ &\mathrm{E}_{\hat{r} \sim G_{\theta}(\hat{r}|q_i)}[\log (1- D_{\phi}(\hat{\varepsilon}|e_{i}))] \\
+ &\mathrm{E}_{r \sim P_{-}(r^{-}|q_i)}[\log (1- D_{\phi}(\varepsilon^{-}|e_{i}))] \}
\end{aligned}
\end{equation}
, where $e_{i}$, $\varepsilon_{i}$, $\hat{\varepsilon}$, and $\varepsilon^{-}$ represent the sentence embeddings of the $i$-th query $q_i$, the positive response $r_i$ paired with $q_i$,  the generative response $\hat{r}$ and the negative retrieval response  $r_{-}$, respectively. $D_{\phi}(\varepsilon_{i}|e_{i})$ represents the matching score calculated by the Discriminator $D_{\phi}$ using Eq.~\ref{score}.

\begin{equation}
\label{discriminator}
\begin{aligned}
s(q_i, r_i)= &D_{\phi}(\varepsilon_{i}|e_{i}) \\
= &\sigma(W_{M} \times concat[f_M(\varepsilon_{i}), f_M(e_{i}), |f_M(\varepsilon_{i})-f_M(e_{i})|]) \\
= &\sigma(W_{M} \times concat[p^{+}, p^{o},|p^{+} - p^{o}|] )
\end{aligned}
\end{equation}

\subsubsection{Details of Generator $G_\theta$}

The hidden state vector $h$ output by T5-Encoder in $D_\phi$ will be taken as input by $G_\theta$, and the T5-Decoder will generate a sequence $\hat{r}_{[1:T]}=w_1, w_2, ..., w_T$. At every timestep, T5-Decoder will predict the probability that the word $w_i$ should be picked out from the vocabulary.

\textbf{Policy Gradient and MC Search}. Given a incomplete sequence $\hat{r}_{[1:t]}$, the unknown last $T-t$ words will be sampled by applying Monte Carlo (MC) search with a roll-out policy. To get a batch of responses, $G_\theta$ run the roll-out policy starting from current timestep till the end of the sequence for $n$ times followed by \cite{seqgan}. Then, $D_\phi$ calculates the scores for these $n$ responses as the reward for the MC search to take the next action (word). The number of responses output by $G_\theta$ can be controlled by the MC search run times $n$.
 
\textbf{Warm-up Training on T5}. Since the T5-Encoder is already trained in $D_\phi$ by retrieval multi-task learning, the T5-Decoder should be trained in advance. Otherwise, the unbalanced performance between $G_\theta$ and $D_\phi$ will eventually leads to the collapse of the adversarial training. Thus, the whole Encoder-Decoder Model (T5) should be warmed up through maximum likelihood estimation (MLE) before the adversarial training. The MLE optimization objective is cross-entropy loss function between the ground truth $r$ and the generated response $\hat{r}$, defined as:

\begin{equation}
\label{mle}
\mathrm{\ell_{ce}(\hat{r};\delta)} = - \sum_{t}^{T} \log( p(w_t| w_{1:t-1}), r; \delta)
\end{equation}
, where $\delta$ represents the parameters of the whole T5 model, which will be updated via MLE. $\log( p(w_t| w_{1:t-1})$ represents the probability of $w_t$ predicted by $G_\theta$ based on the previous $t-1$ words in sequence $\hat{r}$.   

\textbf{Adversarial Objection of $G_\theta$ }. After warming up, the adversarial training will be applied between $G_\theta$ and $D_\phi$. The MLE optimization Objection only focus on the similarity between the ground-truth $r$ and the generated response $\hat{r}$, while the matching score between $\hat{r}$ and the query $q$ paired with $r$ is not considered. Thus, the matching score calculated by $D_phi$ is used for policy learning as a reward. Followed by \cite{thank-bart}, the policy gradient is defined as:

\begin{equation}
\label{policy}
\nabla_\rho J(\rho)=E\left[R \cdot \nabla_\rho \log \left(P\left(\boldsymbol{w_t} \mid \boldsymbol{q} ; \rho\right)\right)\right]
\end{equation}
, where $\rho$ represents the parameters of the whole T5 model, $R$ represents the matching reward calculated by Eq.~\ref{score}, and $w_t$ is sampled from the distribution of T5 outputs at each decoding time step $t$. The adversarial objectives for $G_\theta$ are the loss of the base model (Eq.~\ref{mle}) and the policy gradient of the matching reward (Eq.~\ref{policy}).

\renewcommand{\algorithmicrequire}{\textbf{Input:}}
\renewcommand{\algorithmicensure}{\textbf{Output:}}
\begin{algorithm}[!hb]
	\caption{Adversarial Training Procedure}
	\label{al}
	\begin{algorithmic}[1]
		\REQUIRE training set ${(q,r)}$;
		\ENSURE well-trained \emph{HeroNet}: T5, $\psi_M$, $\psi_D$ 
		\STATE initial the parameters $\theta$ in Discriminator $G_\theta$
		\STATE /*\textbf{step(i)}: warm-up training:*/
		\FOR{Each warm-up training step of T5}
		    \STATE sample a query $q_i$ from the training set 
    		\STATE input $q_i$ to T5-Encoder and then get the hidden vector $h$
    		\STATE input $h$ to T5-Decoder, and then generate a response $\hat{r}$ 
            \STATE update parameters of T5 with MLE, optimized by Eq.~\ref{mle}  
		\ENDFOR
		\STATE /*\textbf{step(ii)}: pre-train T5-Encoder, $\psi_M$ and $\psi_D$*/
		\FOR{Each retrieval multi-task training step}
		    \STATE sample a query $q_i$ from the training set 
    		\STATE extracted negative queries $q^{-}$ via BM25 method
    		\STATE compute the distance between $q_i$ and $q^{-}$ as Eq.~\ref{distance}  
    		\STATE update parameters of T5-Encoder and $\psi_D$ with MLE
    		\STATE extracted $m$ similar queries of $q_i$ via $\psi_D$
    		\STATE get the $m$ responses $r^{-}_{[1:m]}$ paired with these queries from the training set
    		\STATE merge $r^{-}_{[1:m]}$ and positive response $r_i$ to a set $\{r^{i}_j\}$, $j=1,2,..,m+1$
    		\STATE compute matching scores on $(q_i, \{r^{i}_j\})$ as Eq.~\ref{score}
            \STATE update parameters of T5-Encoder and $\psi_M$ with MLE 
		\ENDFOR
		\STATE /*\textbf{step(iii)}: global train \emph{HeroNet} */
		\FOR{Each training step of Generator}
		    \STATE sample $q_i$ from the training set 
    		\STATE given $q_i$, generate $n$ $\hat{r}_{[1:n]}$ by $G_\theta$ with Monte Carlo Search
            \STATE compute matching scores on $(q_i, \hat{r}_{[1:n]})$ by $D_\phi$ as Eq.~\ref{score}
            \STATE update $G_\theta$ with cross-entropy loss (Eq.~\ref{mle}) and the policy gradient (Eq.~\ref{policy})
		\ENDFOR
		\FOR{Each training step of Discriminator}
		    \STATE sample a query $q_i$ from the training set
		    \STATE generate $n$ negative responses $r^{-}_{[1:n]}$ by $G_\theta$ 
		    \STATE retrieve $m$ negative responses $r^{-}_{[1:m]}$ by trained T5-Encoder and $\psi_M$ in (ii)
    		\STATE merge $r^{-}_{[1:n]}$, $r^{-}_{[1:m]}$ and positive response $r_i$ to a set $\{r^{i}_j\}$, $j=1,2,..,m+n+1$
            \STATE compute match scores on $(q_i, \{r^{i}_j\})$ by $D_\phi$ as Eq.~\ref{score}
            \STATE update $D_\phi$ as Eq.~\ref{hl}
		\ENDFOR
		\RETURN \emph{HeroNet}
	\end{algorithmic}
\end{algorithm}

\textbf{Prior Knowledge-guide Strategy}. The retrieved responses extracted by two adaptors of T5-Encoder well-trained on the retrieval multi-task learning will be scored by $D_\phi$ and then the best one of these responses will be spliced with the user query into a piece of input text as the prior knowledge to guide $G_\theta$. 

\subsubsection{Details of Discriminator $D_\phi$} Given a a query $q^+$ and $m+n+1$ responses set ${r_i^o}$, $i=1,2,...m+n+1$, $D_\phi$ will find the truth response $r^+$ from the set by calculating matching score. The responses set consists of $m$ responses extracted by the retrieval module (T5-Encoder and SQD-specific adaptor $\psi_D$), $n$ responses output by the Generator ($G_\theta$) and truth response $r^+$.  

The aims of $D_\phi$ is to minimize the matching score of negative response ($r^{-}$) and maximize the matching score of positive response ($r^{+}$). Thus, the optimization objective of $D_\phi$ is Hinge Loss function, defined as:

\begin{equation}
\label{hl}
\begin{aligned}
\ell_{h} =&\sum_{i=1}^N \{\max (0, \grave{o}_1-s(q_i, r^{+})+s(q_i,r^{-}_{[1:m]})) \\
+ &\max (0, \grave{o}_2-s(q_i, r^{+})+s(q_i, r^{-}_{[1:n]}))\} +\lambda\|\Theta\|_2^2
\end{aligned}
\end{equation}
, where $r^{-}_{[1:m]}$ represents the $m$ retrieval responses, $\grave{o}_1$ represents the average margin between $r^{+}$ and retrieval responses, $r^{-}_{[1:n]}$ represents the $n$ generated responses, $\grave{o}_1$ represents the average margin between $r^{+}$ and generated responses. Moreover, $\|\Theta\|_2^2$ represents the L2 regularization term and $\lambda$ represents the regularization coefficient, which are used to deal with the overfitting problem.

\subsubsection{Adversarial Training Process}
The training procedure of \emph{HeroNet} is shown by Algorithm \ref{al}: (i) The whole T5 module will be warmed up with MLE, (ii) the T5-Encoder and task-specific adaptors ($\psi_M$ and $\psi_D$) is trained with MLE,  and (iii) \emph{HeroNet} will be updated globally by alternately training $D_\phi$ and $G_\theta$. 

\subsection{Re-rank Learning}
The discriminator $D_\phi$ can be used as a re-ranker after adversarial training with Algorithm \ref{al}. The aim of the re-ranker is to score the matching degree of the $m$ retrieved responses and the $n$ generated responses with the given query and maximize the score of the true response as much as possible. Since the T5-Encoder of $D_\phi$ has already gone through the Retrieval Multi-task Learning and the Generation Adversarial Learning, it will be frozen in the Re-rank Learning cause of it's good performance and the training cost. Only the QRM-specific adaptor $\psi_M$ will be updated with MLE. Moreover, the batch of responses retrieved by BM25 are added to the negative samples for the diversity of negative samples.

\section{Experimental Setup} 
\label{Experimental Setup}

In this section, we will elaborate on the details of the experimental setup. To reveal the effectiveness of the proposed \emph{HeroNet}, we intend to answer the following critical Research Questions (RQs) and the related experimental results will be reported in Section.\ref{Results}. 

\textbf{RQ1.}How does \emph{HeroNet} perform when compared with the other models on Generation and Retrieval Task?

\textbf{RQ2.}How much contribution does the retrieval and generation modules respectively provide, i.e., How do different hyper parameter settings (e.g. the number of retrieval queries $m$, the number of generative responses $n$) affect the performance of \emph{HeroNet}?

\textbf{RQ3.} How much contribution does the three strategies (i.e. the multi-task learning strategy, the loss fusion strategy and the prior knowledge-guide strategy) respectively provide in \emph{HeroNet}?


\subsection{Datasets} 

The real-world large-scale dataset for dialogue generation and retrieve extracted from the Ubuntu IRC channel, namely \textit{Ubuntu}\cite{dataset} is used in our experiments. In \textit{Ubuntu}, the context of each dialog contains more than 3 turns which occurred between two participants ($p_1$ and $p_2$) and the next turn of $p_2$ should be directly generated or selected from the given candidate utterances pool of 120k. To alleviate the information loss caused by the truncation of sentences which are overlength, the utterances in the context will be spliced into a piece of text in reverse order.

\subsection{Comparison Models}

To show the advantages of \emph{HeroNet}, it be compared with the following models in our experiments:
\begin{itemize}
    \item[-] BM25: a ranking function used by search engines to estimate the relevance of documents to a given search query, which is also used in the retrieval-based dialogue systems\cite{BM25}.
    \item[-] S-BERT: a sentence embedding model based on \cite{bert}, which is a siamese network\cite{st5}.
    \item[-] ST5: a sentence embedding model based on pre-trained encoder-decoder model by using MEAN-strategy, which is a siamese network\cite{st5}.
    \item[-] Bi-LSTM: bi-directional LSTM, can better capture the two-way semantic dependence in sentence\cite{LSTM}.
    \item[-] BART: a denoising autoencoder for pre-training sequence-to-sequence models, which is particularly effective when fine tuned for text generation \cite{BART}.
    \item[-] T5: a Text-to-Text Transfer Transformer pre-trained model which can be used in a wide variety of English-based NLP problems (e.g. question answering, document summarization, and sentiment classification) \cite{T5}.
    \item[-] REAT: a Retrieval-Enhanced Adversarial Training method for neural response generation in dialogue systems \cite{REAT}.
    \item[-] Multi-Seq2Seq: a novel ensemble of retrieval-based and generation-based open-domain conversation systems \cite{ma1}.
    
\end{itemize}

\subsection{Evaluation Metrics}

For the task of retrieval, Hit@k, Mean Reciprocal Rank (MRR) and Accuracy (Acc) are used as evaluation metrics in the experiments, where $k$ takes values in 5, 10, 50, respectively. For the task of generation, three evaluation metrics are used in the experiments: Bilingual Evaluation Understudy (BLEU)\cite{bleu}, Recall-Oriented Understudy for Gisting Evaluation (ROUGE-L)\cite{ROUGE}, Metric for Evaluation of Translation with Explicit Ordering (METEOR)\cite{METEOR} and Character n-gram F-score (CHRF) \cite{chrf}.

\subsection{Detail Settings}

We implement all experiments on a server with a 10-core Intel Xeon(R)  
64G CPU and a NVIDIA Tesla-V100 32G GPU. For all neural networks, we optimize them with Adam\cite{Adam}, the max sentence length (\textit{max\_seq\_len}) is set as 256 and the batch size ($bs$) is set as 64. In the warm-up training of the whole T5 module in \emph{HeroNet}, the epoch is set as $5$ and the learning rate ($lr$) is set as $4e$-$4$. In the 10 epochs multi-task training of the T5-Encoder and adaptors ($\psi_D$, $\psi_M$), $lr$ is set as $1e$-$4$ and the negative samples are generated by BM25. In the 20 epochs of adversarial training, $lr$ of $G_\theta$ is set as $2e$-$4$ and $lr$ of $D_\phi$ is set as $1e$-$4$. 


\section{Results Analysis} 
\label{Results}
\subsection{Performance on Generation\&Retrieval Tasks (RQ1)}
The generation performance of each models is shown in Table.~\ref{per-generation}. Compared with generation-based models of seq2seq architecture such as Bi-LSTM, BART-base and T5, \emph{HeroNet} achieves better performance on three metrics: BLEU, ROUGE-L and METEOR. This proves the effectiveness of \emph{HeroNet} on generation task. Taking the best response retrieved by BM25 as prior knowledge and inputting it into T5 together with the user query (Method.\textbf{T5+kg} in Table.~\ref{per-generation}) achieves better performance, which proves that the knowledge-guided strategy is feasible. Compared with hybrid models that assemble generation-based and retrieval-based methods such as REAT and Multi-Seq2Seq, \emph{HeroNet} is also able to achieve significant performance gains. This proves the simple architecture of \emph{HeroNet} is more efficient than the complex architecture of the other ensemble models. 


\begin{table}[htbp]
  \centering
  \caption{Performances on Generation Task}
    \begin{tabular}{ccccc}
    \hline
    Method & BLEU  & ROUGE-L & METEOR & CHRF \\
    \hline
    Bi-LSTM & 0.99  & 8.44  & 0.0438 & 11.62 \\
    BART-base  & 1.02  & 8.87  & 0.0445 & 11.76 \\
    T5    & 1.22  & 7.95  & 0.0517 & 12.38 \\
    T5+kg & 2.85  & 9.31  & 0.0525 & 14.11 \\
    REAT  & 3.13  & 8.62  & 0.0616 & 13.22 \\
    Multi-Seq2Seq & 4.24  & 7.80   & 0.0557 & 13.58 \\
    \textbf{\emph{HeroNet}} & \textbf{8.18} & \textbf{11.73} & \textbf{0.0910} & \textbf{16.65} \\
    \hline
    \end{tabular}%
  \label{per-generation}
\end{table}%

The retrieval performance of each models is shown in Table.~\ref{per-retrieval}. Compared with the classic retrieval algorithm BM25, \emph{HeroNet} achieves better performance on three metrics that measures the quality of response ranking: Hit@k, MRR and Acc. This proves the effectiveness of \emph{HeroNet} on retrieval task. Compared with the modish sentence embedding model (ST5 and S-BERT) with siamese architecture, \emph{HeroNet} also achieves better performance. Basically, the retrieval module of \emph{HeroNet} is a kind of st5-like model, and its significant performance improvement compared to ST5 proves the necessity of adversarial training between retrieval module and generation module in \emph{HeroNet}.


\begin{table}[htbp]
  \centering
  \caption{Performances on Retrieval Task}
    \begin{tabular}{cccccc}
    \hline
    Method & MRR   & Acc   & Hit@5 & Hit@10 & Hit@50 \\
    \hline
    BM25  & 0.0314 & 0.0315 & 0.0613 & 0.0831 & 0.1371 \\
    S-BERT & 0.0411 & 0.0392 & 0.0765 & 0.0112 & 0.1582 \\
    ST5   & 0.0346 & 0.0410 & 0.0638 & 0.0924 & 0.1416 \\
    \textbf{\emph{HeroNet}} & \textbf{0.0679} & \textbf{0.0458} & \textbf{0.0937} & \textbf{0.1196} & \textbf{0.1791} \\
    \hline
    \end{tabular}%
  \label{per-retrieval}
\end{table}%

\subsection{Effect of Generation\&Retrieval Tasks on Each Other (RQ2)}

\begin{figure*}[!h]
    \centering                                                  
    \subfigure[BLEU]{%
        \begin{minipage}[t]{0.3\columnwidth}
        \centering                                                 
        \includegraphics[width=\columnwidth]{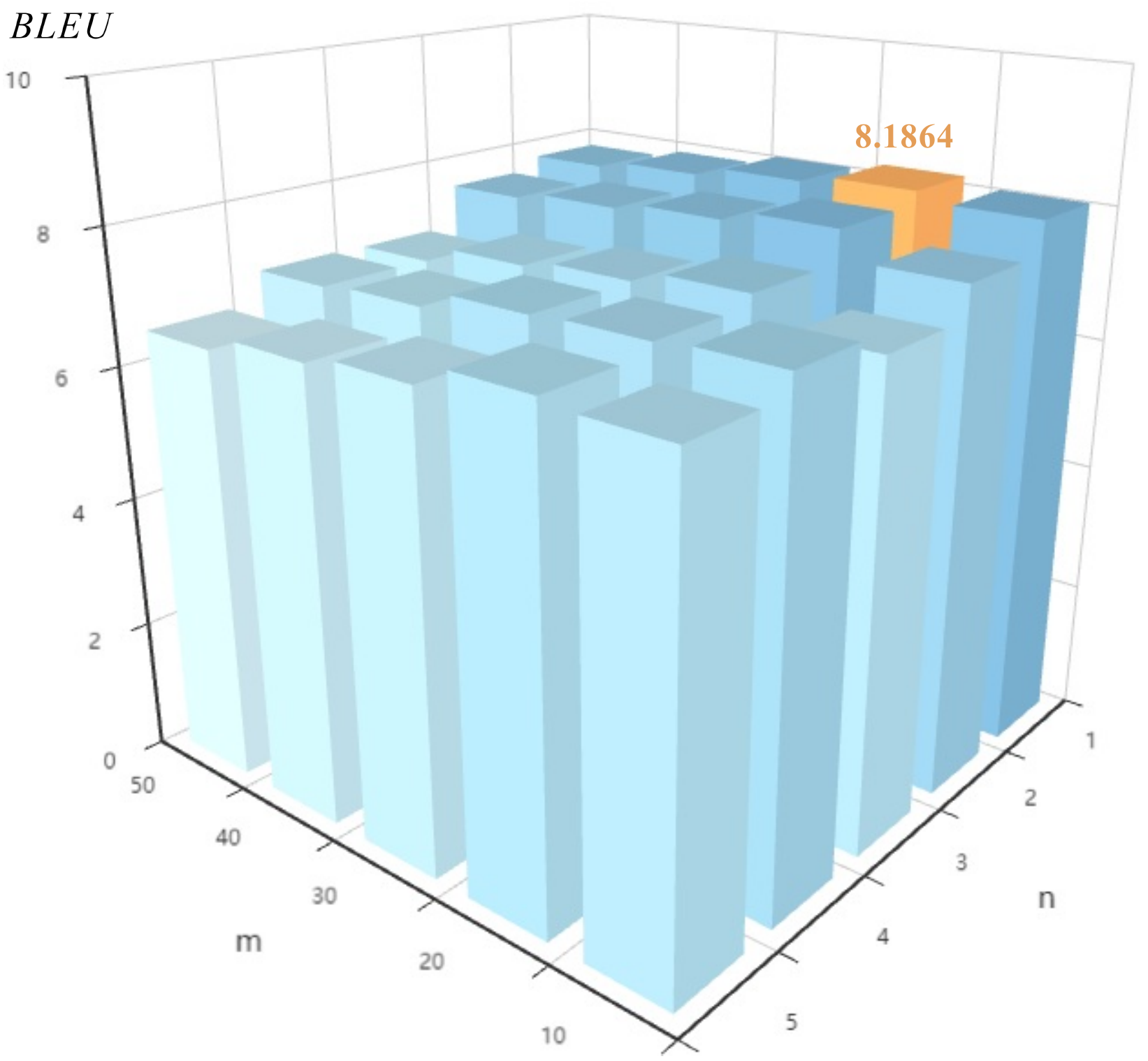} 
        \end{minipage}
        \label{figag}
    }
    \subfigure[ROUGE-L]{%
        \begin{minipage}[t]{0.3\columnwidth}
        \centering                                                 
        \includegraphics[width=\columnwidth]{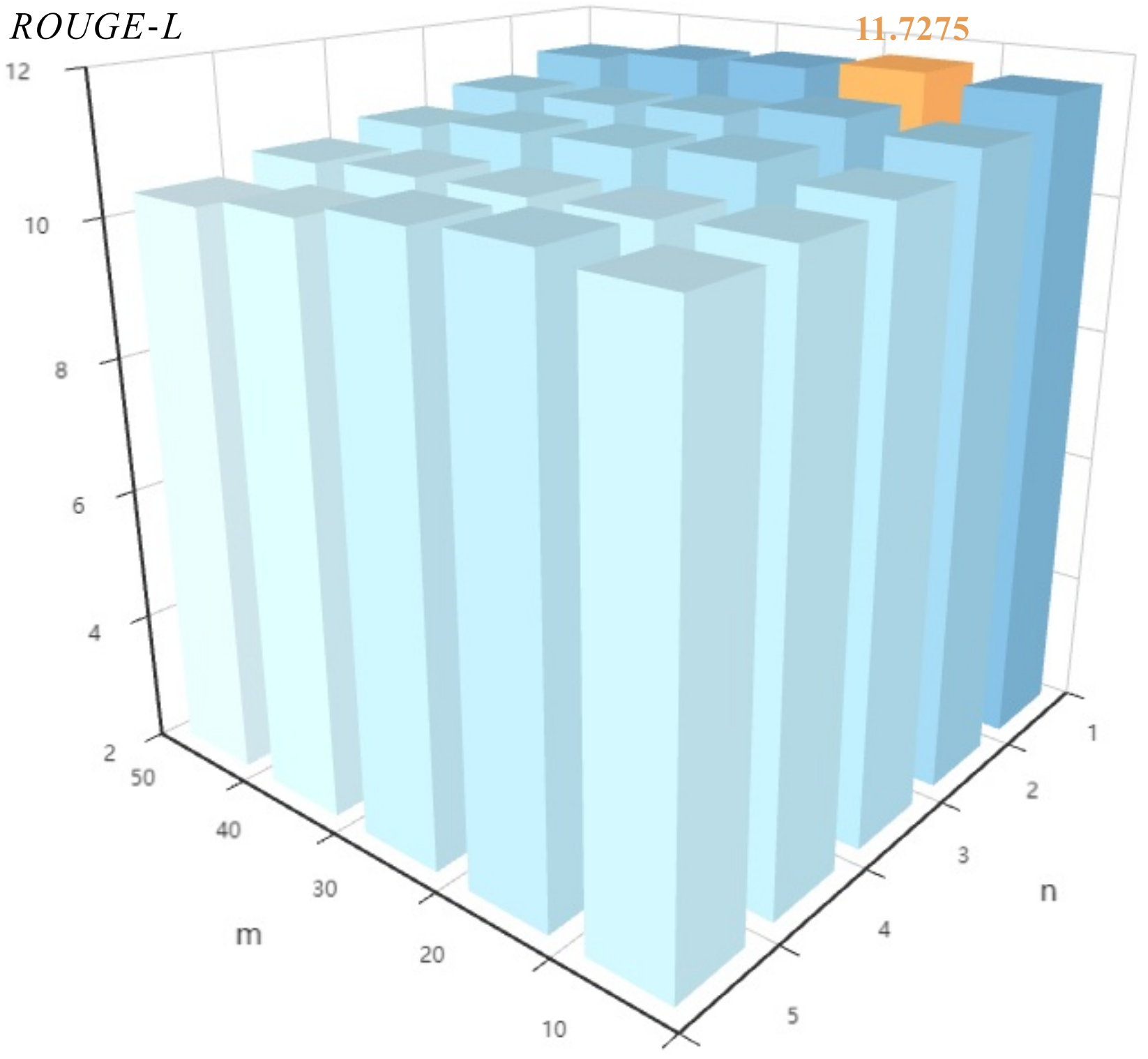}            
        \end{minipage}
        \label{figbg}
    }
    \subfigure[METEOR]{%
        \begin{minipage}[t]{0.3\columnwidth}
        \centering                                                
        \includegraphics[width=\columnwidth]{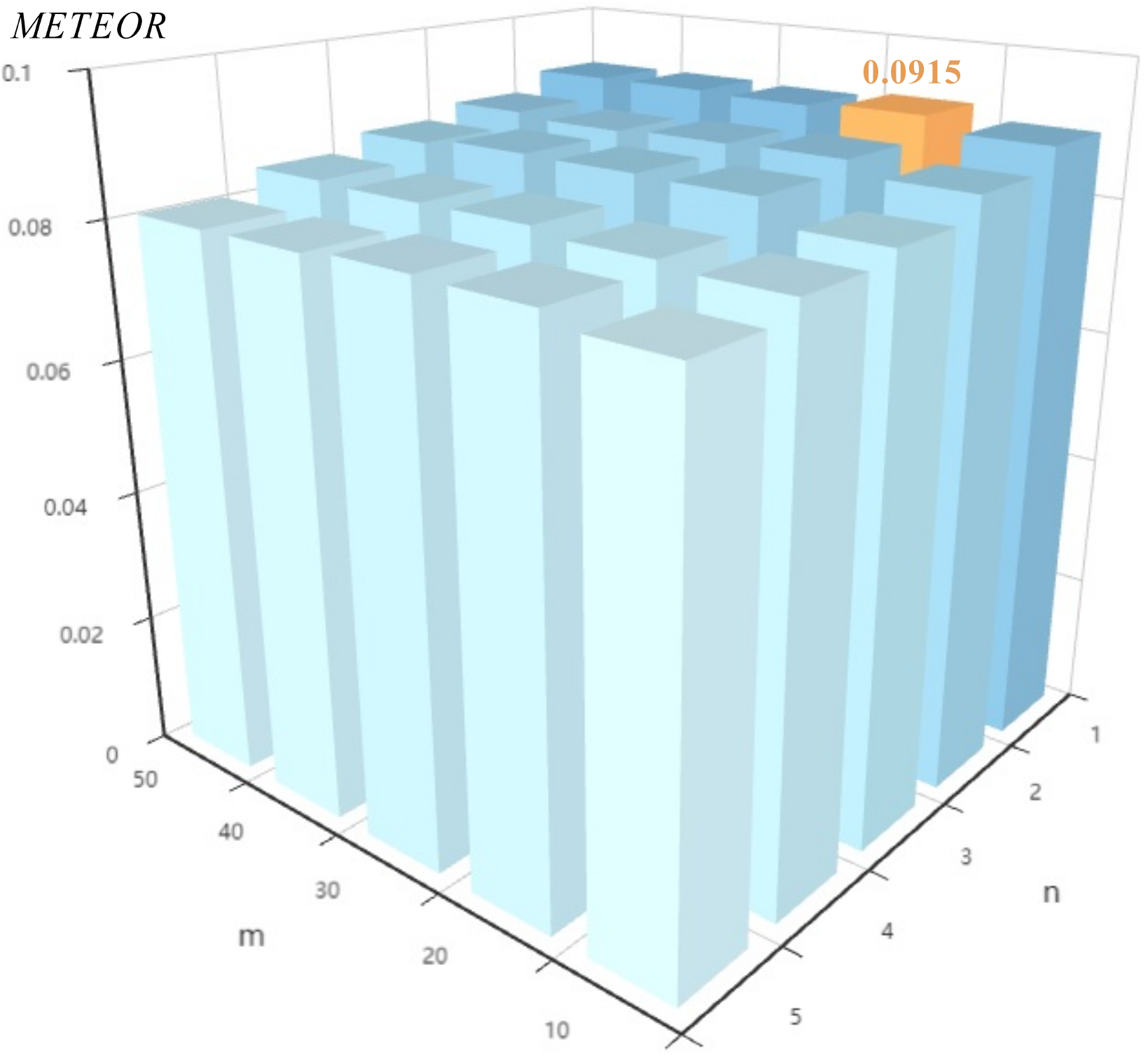}          
        \end{minipage}
        \label{figcg}
    }
    \caption{The performances of \emph{HeroNet} on generation task with different $m$ and $n$.}
    \label{3dg}                                  
\end{figure*}

Fig.~\ref{3dg} shows the generative performances of \emph{HeroNet} when $m$ and $n$ take different values, where $n$ represents the number of responses generated by $G_\theta$ and $m$ represents the number of responses retrieved by $\psi_D$, $\psi_M$ and BM25. The performance of \emph{HeroNet} on generation task shows an upward trend as $n$ decreases when $m$ is fixed. This is because the quality of responses generated by $G_\theta$ will gradually improve as n decreases and these high-quality responses provide better candidates for \emph{HeroNet} in the re-rank learning stage, resulting in an improvement in the overall performance of \emph{HeroNet}.  Since the quality of the top-ranked objects in all retrieved responses is better, the noise will be introduced in the learning process as $m$ increases. Thus, the performance of \emph{HeroNet} shows a downward trend as $m$ increases when $n$ is fixed. In other words, the contribution of the retrieval module to the performance improvement on the generation task is smaller than that of the generation module. When $m$ is set to 20 and $n$ is set to 1, the best performance peaks at 8.1864, 11.7275 and 0.0915 on BLEU, ROUGE-L, and METEOR, respectively. 

\begin{figure*}[!h]
    \centering                                                  
    \subfigure[MRR]{%
        \begin{minipage}[t]{0.3\columnwidth}
        \centering                                                 
        \includegraphics[width=\columnwidth]{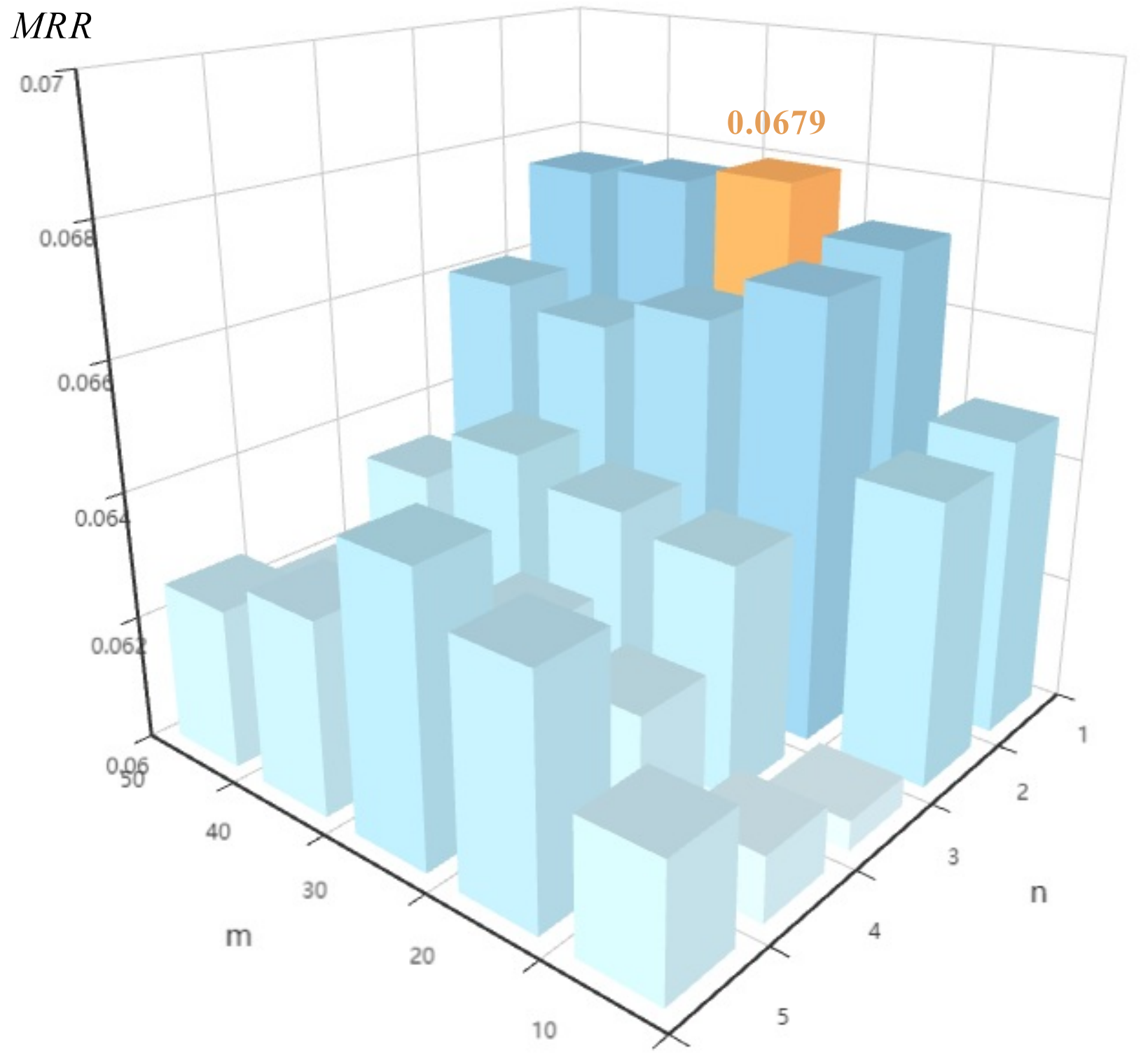} 
        \end{minipage}
        \label{figar}
    }
    \subfigure[Acc]{%
        \begin{minipage}[t]{0.3\columnwidth}
        \centering                                                 
        \includegraphics[width=\columnwidth]{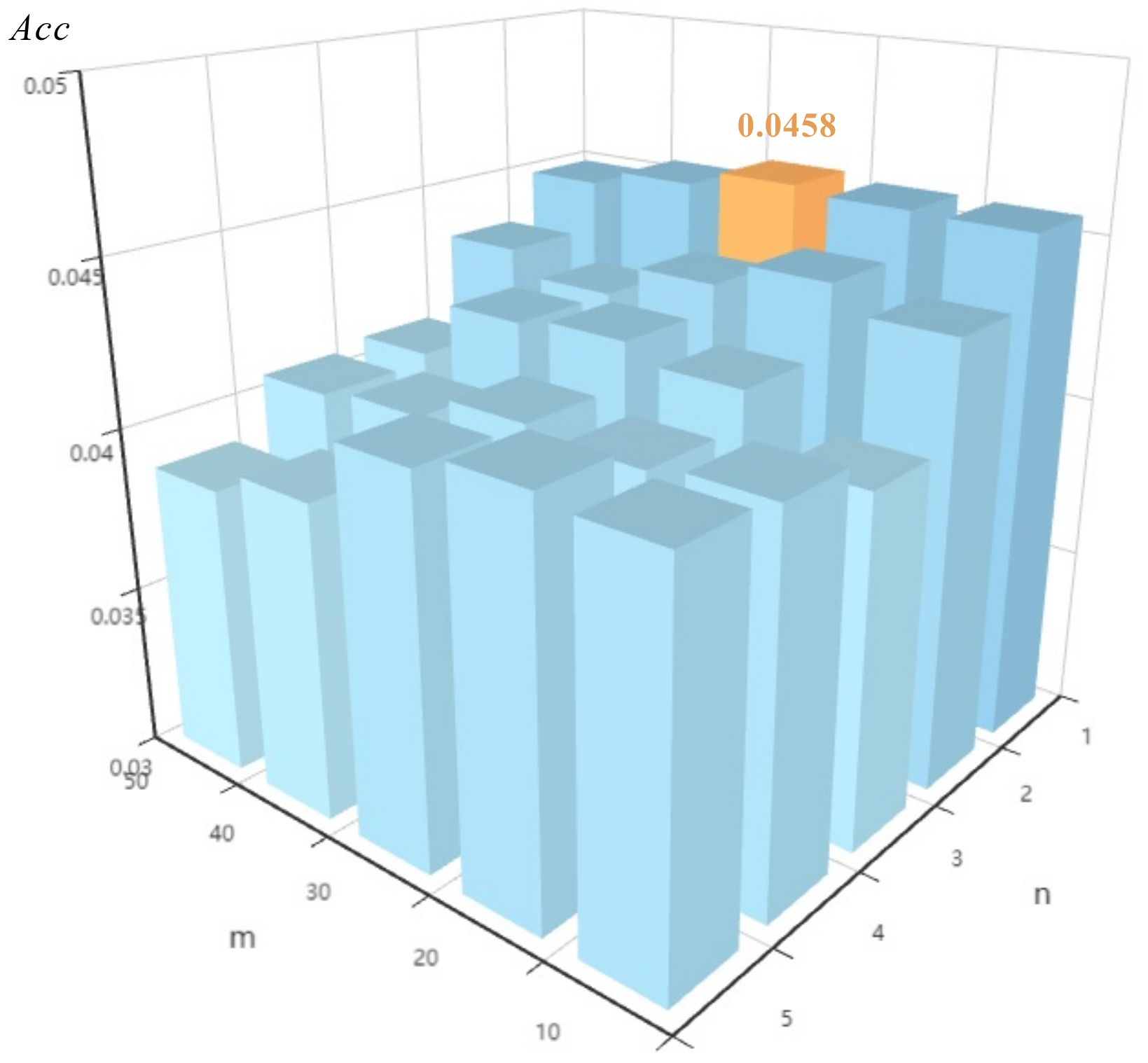}            
        \end{minipage}
        \label{figbr}
    }
    \subfigure[Hit@50]{%
        \begin{minipage}[t]{0.3\columnwidth}
        \centering                                                
        \includegraphics[width=\columnwidth]{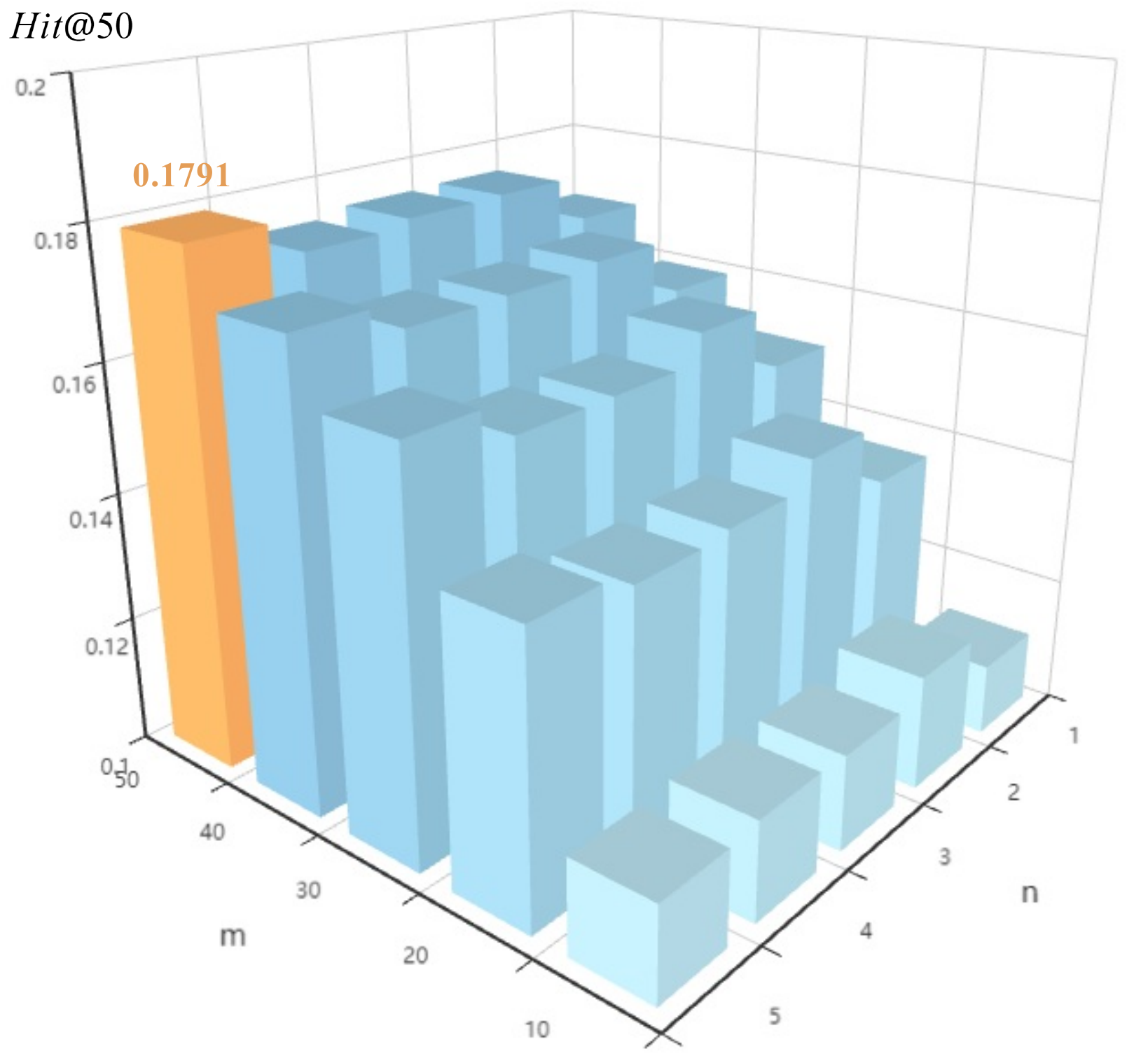}          
        \end{minipage}
        \label{figcr}
    }
    \caption{The performances of \emph{HeroNet} on retrieval task with different $m$ and $n$.}
    \label{3dr}                                  
\end{figure*}

When $m$ and $n$ take different values, the retrieval performance of \emph{HeroNet} on three metrics MRR, Acc, and Hit@50 is shown in Fig.~\ref{3dr}. The improvement trend of retrieval performance on metrics MRR and Acc is roughly the same as that of generation performance while the trend of performance on the metric Hit@50 is just opposite to that of generation performance. As $n$ increases, these high-quality responses generated by $G_\theta$ may be more in line with the user's query than the ground truth, and $D_\phi$ will tend to rank them higher and the ground truth lower. As $m$ increases, the probability of the ground truth appearing in candidate sequences gradually increases, which will be re-ranked together with the generated responses. Simply put, the contribution of the generation module to the performance improvement on the retrieval task is smaller than that of the retrieval module. When $m$ is set to 30 and $n$ is set to 1, the best performance peaks at 0.0678 and 0.0458  on MRR and Acc, respectively. The best performance of Hit@50 peaks at 0.1791, when $m$ is set to 50 and $n$ is set to 5.

\subsection{Ablation Study (RQ3)}

\begin{figure*}[!h]
    \centering                                                  
    \subfigure[Acc]{%
        \begin{minipage}[t]{0.3\columnwidth}
        \centering                                                 
        \includegraphics[width=\columnwidth]{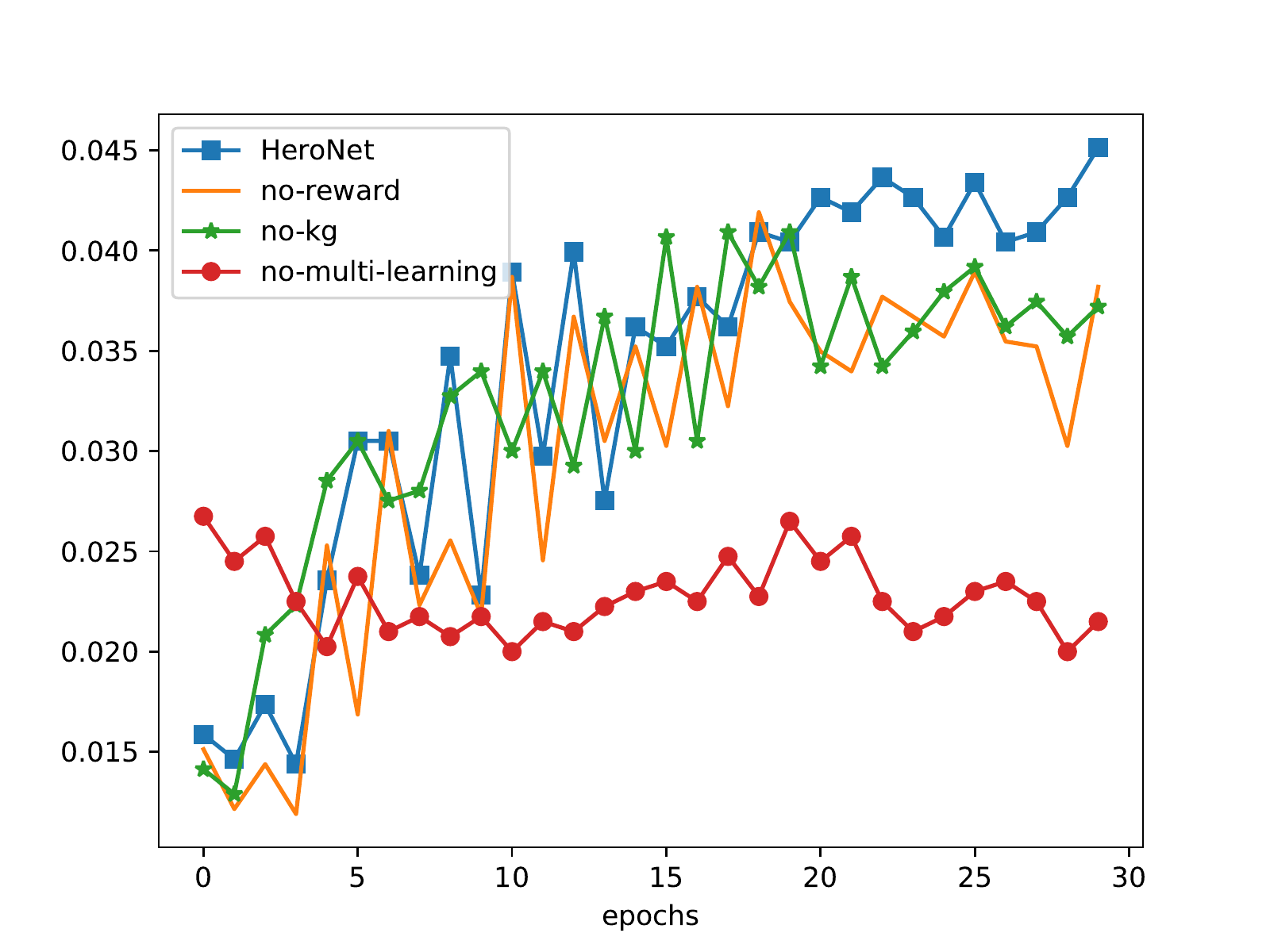} 
        \end{minipage}
        \label{figa}
    }
    \subfigure[BLEU]{%
        \begin{minipage}[t]{0.3\columnwidth}
        \centering                                                 
        \includegraphics[width=\columnwidth]{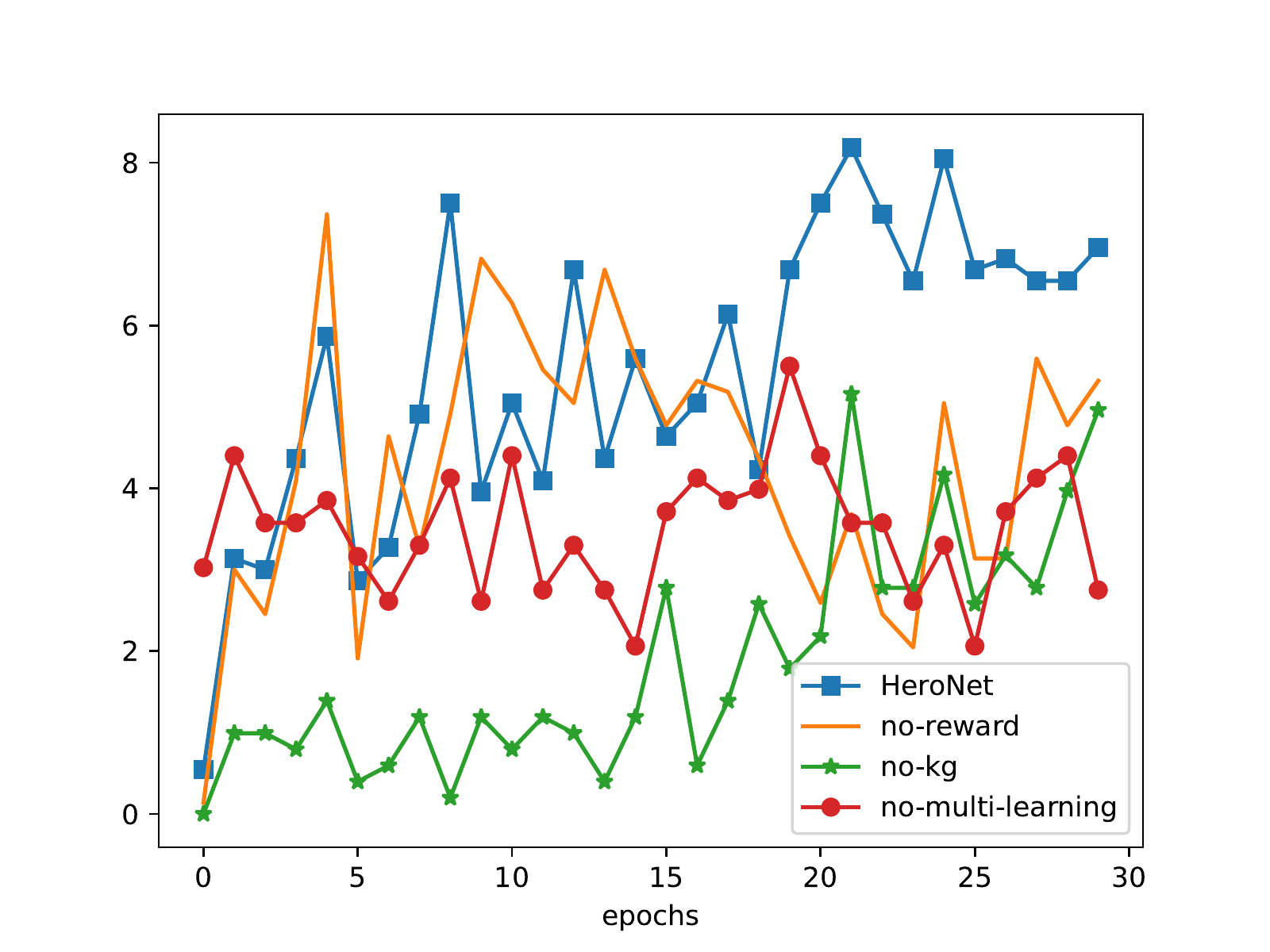}            
        \end{minipage}
        \label{figb}
    }
    \subfigure[Loss]{%
        \begin{minipage}[t]{0.3\columnwidth}
        \centering                                                
        \includegraphics[width=\columnwidth]{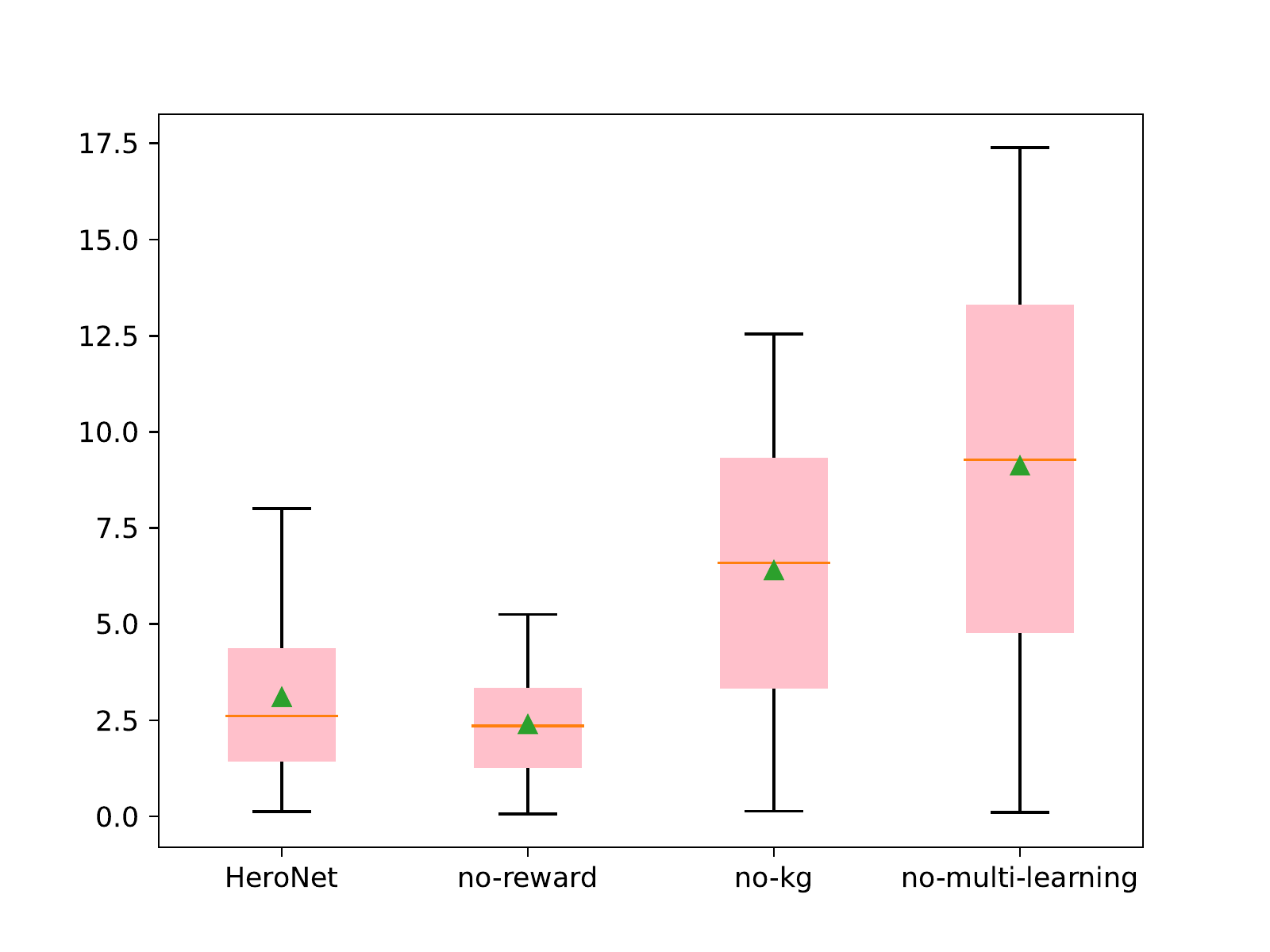}          
        \end{minipage}
        \label{figc}
    }
    \caption{The best performances of different variants of \emph{HeroNet}. \ref{figa} shows the accuracy of these variants on each training epoch, \ref{figb} shows the matrix BLUE of the variants on each epoch, and \ref{figc} shows the spread of training losses of the variants. }
    \label{ablation}                                  
\end{figure*}

To figure out what impact each part of the \emph{HeroNet} has on the overall performance, the following three operations need to be performed on the \emph{HeroNet}: i) not to input retrieval results as prior knowledge to $G_\theta$ (\textbf{no-kg}), ii) not to use the scores output by $D_\phi$ as the reward for the policy gradient (\textbf{no-reward}), and iii) not to apply multi-task learning on T5-Encoder but use two T5-Encoders that share different parameters (\textbf{no-multi-learning}). As shown in Fig.~\ref{figa}, \emph{HeroNet} without multi-task learning achieves the worst performance on Acc, which means that multi-task learning is crucial for the retrieval task. As shown in Fig.~\ref{figb}, without prior knowledge, \emph{HeroNet} achieves the worst performance on the matrix BLEU before 20 epochs, which means that the knowledge-guide strategy has a large impact on the generation task. After 20 epochs, \emph{HeroNet} achieves the worst performance on the matrix BLEU without multi-task learning strategy, which means this strategy has a more important impact than the prior knowledge-guide strategy on the final generative performance of \emph{HeroNet}. As shown in Fig.~\ref{figc}, without the loss fusion strategy, the loss fluctuation of \emph{HeroNet} at each epoch is minimal. This means that the introduction of the policy strategy makes the training unstable, although it eventually achieves the best results in the end. Each variant of \emph{HeroNet} shown in Figure.~\ref{ablation} is the best performing model on the three metrics Acc, BLEU and Loss. 

When $n$ is fix to 1 and $m$ is set to 20 or 50, Table.~\ref{tab:ablation} shows the more detailed results of ablation experiments. In this case, it is obvious that the impact of the three operations on \emph{HeroNet} is still consistent with the above analysis results. Without the multi-task learning strategy, the performance of \emph{HeroNet} bottoms out on the metrics BLEU, ROUGE-L, METEOR, MRR, Hit@5 and Hit@10, again demonstrating the effectiveness of the strategy. Maybe this is  because the encoder after multi-task training has a stronger ability of sentence representation, compared 
to the strategy of training different encoders on each task. Thus, the strategy of multi-task learning may be the key to the effectiveness of \emph{HeroNet} with such a simple architecture.

\begin{table}[!h]
  \centering
  \caption{Ablation Experiment Results of \emph{HeroNet}}
    \begin{tabular}{c|c|cccc|ccccc}
    \toprule
    \multicolumn{1}{c|}{\multirow{2}[4]{*}{\shortstack{Methods \\ $n$=1}}} & \multirow{2}[4]{*}{$m$} & \multicolumn{4}{c|}{Generation Task} & \multicolumn{5}{c}{Retrieval Task} \\
\cmidrule{3-11}  &  & BLEU  & ROUGE-L & METEOR & CHRF  & MRR   & Acc   & Hit-5 & Hit-10 & Hit-50 \\   
    \hline
    \multirow{2}[2]{*}{\emph{HeroNet}} & 20    & 8.18  & 11.64 & 0.0909 & 16.55 & 0.0672 & 0.0458 & 0.0927 & 0.1166 & 0.1360 \\
          & 50    & 7.79  & 11.41 & 0.0911 & 16.28 & 0.0675 & 0.0448 & 0.0914 & 0.1174 & 0.1683 \\
    \midrule
    \multirow{2}[2]{*}{\shortstack{no- \\ kg}} & 20    & 7.25  & 11.18 & 0.0856 & 16.22 & 0.0624 & 0.039 & 0.0907 & 0.1138 & \underline{0.1317} \\
          & 50    & 6.66  & 10.91 & 0.0838 & 15.88 & 0.0611 & 0.0355 & 0.0907 & 0.1133 & 0.1659 \\
    \midrule
    \multirow{2}[2]{*}{\shortstack{no- \\ reward}} & 20    & 7.54  & 11.39 & 0.0928 & 16.02 & 0.0669 & 0.0461 & 0.0929 & 0.1143 & 0.1342 \\
          & 50    & 6.93  & 11.01 & 0.0915 & \underline{15.43} & 0.0659 & 0.0421 & 0.0894 & 0.1173 & 0.1672 \\
    \midrule
    \multirow{2}[2]{*}{\shortstack{no-multi\\learning}} & 20    & 6.49  & 9.95  & 0.0672 & 16.33 & 0.055 & 0.0327 & 0.0818 & 0.1065 & 0.1322 \\
          & 50    & \underline{4.98}  & \underline{9.39}  & \underline{0.0613} & 16.12 & \underline{0.049} & \underline{0.0272} & \underline{0.0690} & \underline{0.1010} & 0.1629 \\
    \bottomrule
    \end{tabular}%
  \label{tab:ablation}%
\end{table}%


\section{Conclusion}
This paper proposes a \textbf{h}ybrid r\textbf{e}t\textbf{r}ieval-generati\textbf{o}n \textbf{net}work (\emph{\textbf{HeroNet}}), which has a simple but effective architecture applied three process of learning. By introducing multi-task learning, the performance of sentence embedding is improved in \emph{HeroNet}. Moreover, the generation and retrieval performance of \emph{HeroNet} is also improved by exploiting adversarial training and prior knowledge. \emph{HeroNet} consists of only one encoder, one decoder and two adapters, this simple architecture makes it easy to train.

%
%
%
\bibliographystyle{splncs04}
\bibliography{ref}

\begin{thebibliography}{10}
\providecommand{\url}[1]{\texttt{#1}}
\providecommand{\urlprefix}{URL }
\providecommand{\doi}[1]{https://doi.org/#1}

\bibitem{METEOR}
Banerjee, S., Lavie, A.: {METEOR}: An automatic metric for {MT} evaluation with
  improved correlation with human judgments. In: Proc. of ACL Workshop (2005)

\bibitem{bert}
Devlin, J., Chang, M., Lee, K., Toutanova, K.: {BERT:} pre-training of deep
  bidirectional transformers for language understanding. In: Proc. of NAACL
  (2019)

\bibitem{MaskGAN}
Fedus, W., Goodfellow, I.J., Dai, A.M.: Maskgan: Better text generation via
  filling in the {\_}{\_}{\_}{\_}{\_}{\_}{\_}. In: Proc. of ICLR (2018)

\bibitem{multi-dataset}
Friedman, D., Dodge, B., Chen, D.: Single-dataset experts for multi-dataset
  question answering. In: Proc. of EMNLP (2021)

\bibitem{gan}
Goodfellow, I.J., Pouget{-}Abadie, J., Mirza, M., Xu, B., Warde{-}Farley, D.,
  Ozair, S., Courville, A.C., Bengio, Y.: Generative adversarial nets. In:
  Proc. of NeurIPS (2014)

\bibitem{LeakedGAN}
Guo, J., Lu, S., Cai, H., Zhang, W., Yu, Y., Wang, J.: Long text generation via
  adversarial training with leaked information. In: Proc. of AAAI (2018)

\bibitem{LSTM}
Hochreiter, S., Schmidhuber, J.: Long short-term memory. Neural Comput.  (1997)

\bibitem{Adam}
Kingma, D.P., Ba, J.: Adam: {A} method for stochastic optimization. In: Proc.
  of ICLR (2015)

\bibitem{thank-bart}
Lai, H., Toral, A., Nissim, M.: Thank you bart! rewarding pre-trained models
  improves formality style transfer. In: Proc. of ACL (2021)

\bibitem{BART}
Lewis, M., Liu, Y., Goyal, N., Ghazvininejad, M., Mohamed, A., Levy, O.,
  Stoyanov, V., Zettlemoyer, L.: {BART:} denoising sequence-to-sequence
  pre-training for natural language generation, translation, and comprehension.
  In: Proc. of ACL (2020)

\bibitem{Rpros}
Li, J., Liu, C., Tao, C., Chan, Z., Zhao, D., Zhang, M., Yan, R.: Dialogue
  history matters! personalized response selection in multi-turn
  retrieval-based chatbots. {ACM} Trans. Inf. Syst.  (2021)

\bibitem{ma3}
Li, X., Liu, J., Zheng, W., Wang, X., Zhu, Y., Dou, Z.: A hybrid framework of
  emotion-aware seq2seq model for emotional conversation generation. In: Proc.
  of NTCIR (2019)

\bibitem{ROUGE}
Lin, C.Y.: {ROUGE}: A package for automatic evaluation of summaries. In: Text
  Summarization Branches Out (2004)

\bibitem{dataset}
Lowe, R., Pow, N., Serban, I., Pineau, J.: The ubuntu dialogue corpus: {A}
  large dataset for research in unstructured multi-turn dialogue systems. In:
  Proc. of SIGDIAL (2015)

\bibitem{st5}
Ni, J., {\'{A}}brego, G.H., Constant, N., Ma, J., Hall, K.B., Cer, D., Yang,
  Y.: Sentence-t5: Scalable sentence encoders from pre-trained text-to-text
  models. In: Proc. of ACL Findings (2022)

\bibitem{RelGAN}
Nie, W., Narodytska, N., Patel, A.: Relgan: Relational generative adversarial
  networks for text generation. In: Proc. of ICLR (2019)

\bibitem{mb1}
Pandey, G., Contractor, D., Kumar, V., Joshi, S.: Exemplar encoder-decoder for
  neural conversation generation. In: Proc. of ACL (2018)

\bibitem{bleu}
Papineni, K., Roukos, S., Ward, T., Zhu, W.: Bleu: a method for automatic
  evaluation of machine translation. In: Proc. of ACL (2002)

\bibitem{chrf}
Popovic, M.: chrf: character n-gram f-score for automatic {MT} evaluation. In:
  In: Proc.of WMT@EMNLP. pp. 392--395 (2015)

\bibitem{T5}
Raffel, C., Shazeer, N., Roberts, A., Lee, K., Narang, S., Matena, M., Zhou,
  Y., Li, W., Liu, P.J.: Exploring the limits of transfer learning with a
  unified text-to-text transformer. J. Mach. Learn. Res.  (2020)

\bibitem{BM25}
Robertson, S.E., Zaragoza, H.: The probabilistic relevance framework: {BM25}
  and beyond. Found. Trends Inf. Retr.  (2009)

\bibitem{ma1}
Song, Y., Li, C., Nie, J., Zhang, M., Zhao, D., Yan, R.: An ensemble of
  retrieval-based and generation-based human-computer conversation systems. In:
  Proc. of IJCAI (2018)

\bibitem{IRGAN}
Wang, J., Yu, L., Zhang, W., Gong, Y., Xu, Y., Wang, B., Zhang, P., Zhang, D.:
  {IRGAN:} {A} minimax game for unifying generative and discriminative
  information retrieval models. In: Proc. of SIGIR (2017)

\bibitem{pg}
Williams, R.J.: Simple statistical gradient-following algorithms for
  connectionist reinforcement learning. Mach. Learn.  (1992)

\bibitem{esbot}
Xu, X., Wang, Z., Tu, Z., Chu, D., Ye, Y.: {E-SBOT:} {A} soft service robot for
  user-centric smart service delivery. In: 2019 IEEE World Congress on
  Services, SERVICES, Milan, Italy, July 8-13, 2019 (2019)

\bibitem{HybridNCM}
Yang, L., Hu, J., Qiu, M., Qu, C., Gao, J., Croft, W.B., Liu, X., Shen, Y.,
  Liu, J.: A hybrid retrieval-generation neural conversation model. In: Proc.
  of CIKM (2019)

\bibitem{seqgan}
Yu, L., Zhang, W., Wang, J., Yu, Y.: Seqgan: Sequence generative adversarial
  nets with policy gradient. In: Proc. of AAAI (2017)

\bibitem{bolin}
Zhang, B., Tu, Z., Jiang, Y., He, S., Chao, G., Chu, D., Xu, X.: {DGPF:} {A}
  dialogue goal planning framework for cognitive service conversational bot.
  In: Proc. of IEEE ICWS (2021)

\bibitem{EnsembleGAN}
Zhang, J., Tao, C., Xu, Z., Xie, Q., Chen, W., Yan, R.: Ensemblegan:
  Adversarial learning for retrieval-generation ensemble model on short-text
  conversation. In: Proc. of SIGIR (2019)

\bibitem{ma2}
Zhang, L., Yang, Y., Zhou, J., Chen, C., He, L.: Retrieval-polished response
  generation for chatbot. {IEEE} Access  (2020)

\bibitem{REAT}
Zhu, Q., Cui, L., Zhang, W., Wei, F., Liu, T.: Retrieval-enhanced adversarial
  training for neural response generation. In: Proc. of ACL (2019)

\end{thebibliography}
\end{document}